\documentclass{article}

\usepackage[preprint,nonatbib]{neurips_2022}

\usepackage[utf8]{inputenc} 
\usepackage[T1]{fontenc}    
\usepackage{url}            
\usepackage{booktabs}       
\usepackage{amsfonts}       
\usepackage{nicefrac}       
\usepackage{microtype}      
\usepackage{xcolor}         
\usepackage{graphicx}
\usepackage{amsmath}
\usepackage{amsthm}
\usepackage{booktabs}
\usepackage[capitalize]{cleveref}
\usepackage{enumitem}
\usepackage{algorithm}
\usepackage[noend]{algpseudocode}
\usepackage{algorithmicx,algorithm}
\usepackage{multirow}
\usepackage{amssymb}
\usepackage{amsfonts}
\usepackage{array}
\usepackage{caption}
\usepackage{subcaption}
\usepackage{makecell}
\usepackage{chngcntr}
\usepackage{apptools}
\usepackage{bm}
\AtAppendix{\counterwithin{lemma}{section}}
\AtAppendix{\counterwithin{theorem}{section}}
\AtAppendix{\counterwithin{prop}{section}}

\newtheorem{theorem}{Theorem}
\newtheorem{lemma}{Lemma}
\newtheorem{prop}{Proposition}

\newtheorem*{definition}{Definition}

\DeclareRobustCommand\onedot{\futurelet\@let@token\@onedot}
\def\eg{\emph{e.g.}} 
\def\ie{\emph{i.e.}}

\def\etal{\emph{et al.}}

\usepackage{xurl}
\title{Towards Explanation for Unsupervised Graph-Level Representation Learning}

\author{%
  Qinghua Zheng\textsuperscript{1}\quad Jihong Wang\textsuperscript{1}\quad Minnan Luo\textsuperscript{1}\quad Yaoliang Yu\textsuperscript{2}\quad Jundong Li\textsuperscript{3}\quad \\ \textbf{Lina Yao\textsuperscript{4}}\quad \textbf{Xiaojun Chang\textsuperscript{5}}\\
    $^1$Xi'an Jiaotong University\\
    $^2$University of Waterloo\\
    $^3$University of Virginia\\
    $^4$University of New South Wales\\
    $^5$University of Technology Sydney\\
    \texttt{qhzheng@xjtu.edu.cn},\quad  \texttt{wang1946456505@stu.xjtu.edu.cn}, \quad \texttt{minnluo@xjtu.edu.cn}\\
     \texttt{yaoliang.yu@uwaterloo.ca},\quad \texttt{jundong@virginia.edu},\quad \texttt{lina.yao@unsw.edu.au}\\ \texttt{xiaojun.chang@uts.edu.au}\\
}

\begin{document}

\maketitle

\begin{abstract}
    Due to the superior performance of Graph Neural Networks (GNNs) in various domains, there is an increasing interest in the GNN explanation problem "\emph{which fraction of the input graph is the most crucial to decide the model's decision?}" Existing explanation methods focus on the supervised settings, \eg, node classification and graph classification, while the explanation for unsupervised graph-level representation learning is still unexplored. The opaqueness of the graph representations may lead to unexpected risks when deployed for high-stake decision-making scenarios. In this paper, we advance the Information Bottleneck principle (IB) to tackle the proposed explanation problem for unsupervised graph representations, which leads to a novel principle, \textit{Unsupervised Subgraph Information Bottleneck} (USIB). We also theoretically analyze the connection between graph representations and explanatory subgraphs on the label space, which reveals that the expressiveness and robustness of representations benefit the fidelity of explanatory subgraphs. Experimental results on both synthetic and real-world datasets demonstrate the superiority of our developed explainer and the validity of our theoretical analysis.
\end{abstract}

\section{Introduction}

Graph Neural Networks (GNNs) have emerged as a promising learning paradigm and demonstrated superior learning performance on different graph learning tasks, such as node classification~\cite{kipf2017semi,velickovic2018graph,liu2020towards,hamilton2017inductive}, graph classification~\cite{ying2018hierarchical,xu2018how}, and link prediction ~\cite{zhang2018link,cai2021line}. Despite their strengths, GNNs are usually treated as black boxes and thus cannot provide human-intelligible explanations~\cite{ying2019gnnexplainer,luo2020parameterized}. Such opaqueness impedes their broad adoption in many decision-critical applications pertaining to fairness, privacy, and safety~\cite{doshi2017towards}. To better understand the working mechanisms of GNNs, researchers focus on the GNN explanation problem recently: \textit{what knowledge does the GNN model extract to make the specific decision?} Specifically, given a graph and a GNN model, explanation methods try to find the intrinsic graph components that are crucial in the procedure of prediction for the target instance. To the best of our knowledge, all existing works~\cite{pope2019explainability,yuan2020explainability,ying2019gnnexplainer,vu2020pgm} only consider explanations in supervised settings, such as graph classification and node classification. More specifically, these works aim to provide trusty explanations on why graphs and nodes are classified into a specific category by GNN models. 
However, the explanation problem for unsupervised graph representation learning is still unexplored. Different from supervised GNNs, unsupervised graph representation learning methods ~\cite{perozzi2014deepwalk,tang2015line,grover2016node2vec,kipf2016variational} embed graph data into low-dimensional representations without any supervision signals. Because of their opaqueness, it is hazardous to employ these representations for high-stake decision-making problems directly. 
In consequence, it is imperative to lift the veil of unsupervised graph representations by answering the following question: \emph{what knowledge do unsupervised GNNs incorporate to generate the graph representations}?





To provide a universal perspective, in this paper, we focus on the explanation for unsupervised graph-level representations\footnote{The explanation for node-level representations can be seen as a particular case by considering the corresponding $k$-hop neighbor subgraphs of each node as a graph instance.}. Specifically, we aim to find explanatory subgraphs most relevant to the representations while incorporating superfluous information as little as possible. 
There are mainly two challenges in the unexplored explanation problem: (1) Different from supervised GNNs which are designed for specific tasks, unsupervised graph representations are usually task-agnostic. Thus, existing explanation methods based on specific learning tasks cannot be directly grafted to explain unsupervised graph representations. (2) It is computational intractably to evaluate the relevance between the subgraphs and the representation vectors since they are located in different spaces, \ie, the non-Euclidean space and Euclidean space, respectively. As a result, an effective and efficient measurement is urgently needed.

\begin{figure}[t] 
\centering 
\includegraphics[width=0.99\textwidth]{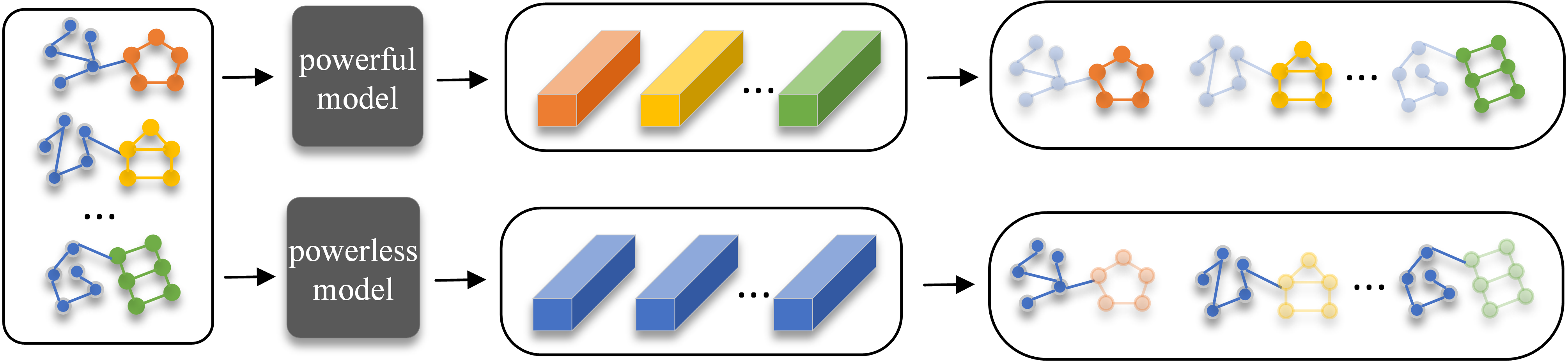} 
\caption{An intuition of explanation for unsupervised graph representation. Given graphs with different motifs (\ie, cycle, house and grid), expressive and robust representations generated by powerful GNNs can exactly extract the motif-related information and ignore the random noises. Thus explanatory subgraphs for the powerful representations can explore the essential structure and vice versa.} 
\label{fig:motivation} 
\end{figure}

To address the above challenges, we propose a novel method named  \textit{Unsupervised Subgraph Information Bottleneck} (USIB) in light of the Information Bottleneck (IB) principle~\cite{yu2021recognizing,wu2020graph} that learns compressed representations from the input data while keeping the most predictive information of labels.
Specifically, USIB adopts the mutual information to formalize the relevance. It maximizes the mutual information between explanatory subgraphs and graph representations while minimizing the mutual information between explanatory subgraphs and given graph instances, simultaneously.  Moreover, we theoretically analyze the connection between representations and explanatory subgraphs on the label space, which reveals that the expressiveness and robustness of representations benefit the fidelity of explanatory subgraphs. 
Intuitively, as shown in~\cref{fig:motivation}, when the graphs have different motifs, expressive and robust representations extract the motif-related information exactly while uninformative and fragile representations focus more on noises. Thus, explanations generated for powerful representations can explore the essential structure, while the fidelity of explanation for powerless representations is not guaranteed.
The contributions of this paper are summarized as follows.
\begin{itemize}[leftmargin=*]
    \item  We consider an unexplored explanation problem. To the best of our knowledge, this is the first work to explore the explanation for unsupervised graph representation learning with GNNs.
    \item 
    We propose a novel explanation method USIB based on the IB principle, which aims to find the most informative yet compressed explanation. We also theoretically analyze the connection between representations and explanatory subgraphs on the label space, revealing that representations' expressiveness and robustness benefit the explanatory subgraphs.
    \item Extensive experiments show that our method achieves the state-of-the-art performance on several benchmark datasets 
    and also demonstrate the validity of our theoretical analysis.
\end{itemize}

\vspace{-0.3cm}
\section{Related works}


\paragraph{Instance-level explanation.} Instance-level explanation aims to give explanations for specific instances. In general, there are mainly two lines of works: (1) Non-parametric explanation methods~\cite{baldassarre2019explainability,pope2019explainability,schnake2020higher,sundararajan2017axiomatic} usually involve gradient-like scores as heuristics to quantify the importance of edges and nodes. The gradient-like scores can be easily obtained by conducting backpropagation from the target model prediction to the input, \eg, adjacency matrix. For example, SA~\cite{baldassarre2019explainability} directly employs the squared values of gradients as the importance scores of different input features. However, these methods can only reflect the sensitivity between input and output, which cannot accurately show the importance. (2) Parametric explanation methods~\cite{ying2019gnnexplainer,schwab2019cxplain,luo2020parameterized,wx2021refine,vu2020pgm} generate the explanatory subgraphs with a parametrized explainer model. To name a few, GNNExplainer~\cite{ying2019gnnexplainer} learns soft masks with a local view and applies the masks on the adjacency matrix. To provide a global understanding of the model prediction, PGExplainer~\cite{luo2020parameterized} generates the explanatory subgraphs with a deep neural network whose parameters are shared across the explained instances. Moreover, to simultaneously exhibit the local and global explainability, Refine~\cite{wx2021refine} adopts the pre-training and fine-tuning techniques to explain GNNs. 

\paragraph{Model-level explanation.} Model-level explanation aims to provide general insights and high-level understanding to explain deep graph models. The only existing model-level method for explaining graph neural networks is XGNN~\cite{yuan2020xgnn} that explains GNNs by training a graph generator, so as to output class-wise graph patterns to explain a specific class. 
However, to our best knowledge, all existing explanation methods focus on end-to-end GNNs in the framework of supervised learning, \eg, node classification and graph classification GNN models are mostly studied. 
In this regard, it is still unexplored how to make explanations for unsupervised graph representation learning.

\section{Notations and preliminaries}
\paragraph{Notations.}In most cases, we denote random variables with upper-case letters (\eg, $G$ and $Z$), and represent its support set by calligraphic letters (\eg, $\mathcal{G}$ and $\mathcal{Z}$). Upper-case letters with superscript (\eg, $G^{(k)}$and $Z^{(m)}$) refers to the instances of random variables correspondingly. 
Suppose the graph data is given as $G=(\mathcal{V},\mathcal{E})$, where $\mathcal{V}$ and $\mathcal{E}$ are the nodes and edges respectively. The representations of graph $G$, denoted as $Z$, is learned by an unsupervised GNN model $f_t:\mathcal{G}\rightarrow\mathcal{Z}$, such as the state-of-the-arts Infograph\cite{sun2019infograph}, GCL~\cite{You2020GraphCL}, and ADGCL~\cite{suresh2021adversarial}. Moreover, we denote the ground-truth labels of graphs as $Y$ which are unreachable in the unsupervised setting. 

\paragraph{Information Bottleneck~\cite{tishby1999information,tishby2015deep}.} Given input data $X$ and its label $Y$, Information Bottleneck aims to discover a compressed latent representation $Z$ that is maximally informative in terms of $Y$. Formally, one can learn the latent representation $Z$ by optimizing the following optimization problem
\begin{equation}
    \max_{Z} \mathcal{L}_{IB} = I(Z;Y) - \beta I(X;Z)
\end{equation}
where $\beta$ denotes a hyper-parameter trading-off the informativeness and compression. Mutual information (MI) $I(X;Z)$ measures the relevance of two random variables, formulated as $I(X;Z) = \int_x\int_z p(x,z)\log \frac{p(x,z)}{p(x)p(z)}dxdz.$

\paragraph{GNN explanation.} 
GNN explanation aims to understand the intrinsic information of the graphs that are crucial for GNN's computation process, so as to provide human-intelligible explanations. Specifically, given a graph $G$ and a GNN model $\psi$ that learns a conditional distribution $P_{\psi}(\hat{Z}|G)$, GNN explanation aims to learn explanatory subgraphs $S$ that are most relevant with GNN's computation results, \ie,
\begin{equation} \label{eq:def}
    S = \arg\mathop{\max}_{S\in\mathcal{S}} Score(S, \hat{Z})
\end{equation}
where $\mathcal{S}$ denotes the universe set consisting of all possible subgraphs of graph $G$; $Score(S,\hat{Z})$ measures the relevance between subgraph $S$ and GNN's computation results $\hat{Z}$. 
For example, GNNExplainer~\cite{ying2019gnnexplainer} focuses on the explanation for supervised GNNs, and formalizes the relevance $Score(S,\hat{Z})$ as mutual information, \ie, $S = \arg\mathop{\max}_{S\in\mathcal{S}} I(S;\hat{Y})$, where random variable $\hat{Y}=\hat{Z}$ refers to the classification probabilities. 

\section{Methodology}
In this section, we first introduce our proposed explanation method for unsupervised graph representation learning, \ie, USIB. Then, we elaborate on the mutual information estimation methods for USIB. Finally, the reparameterization trick is presented for the optimization of USIB objective.

\subsection{Unsupervised Subgraph Information Bottleneck}
In this paper, we study the unexplored explanation problem for unsupervised graph-level representation learning. Given a graph $G$ and its corresponding representation $Z$ extracted by unsupervised GNNs, our goal is to identify the explanatory subgraph $S$ that is most relevant to the representations. Following the previous explanation works~\cite{ying2019gnnexplainer,luo2020parameterized}, we leverage mutual information to measure the relevance and therefore formulate the explanation problem as $\arg\mathop{\max}_S I(S;Z)$. Unfortunately, it has been proved that there is a trivial solution $S = G$ since $I(Z;S)\leq I(Z;G)$ (The proof is shown in the Appendix B).
The trivial solution indicates that the explanatory subgraph $S$ may incorporate superfluous information, \eg, noise and irrelevant information with representations $Z$. Inspired by the success of IB principle in explanation for supervised GNNs~\cite{yu2021recognizing}, we generalize the IB principle to the unsupervised setting to avoid the trivial solution and exploit a novel principle. 
\begin{definition}
(Unsupervised Subgraph Information Bottleneck: USIB). Given a graph $G$ and its representation $Z$, the USIB seeks for the most informative yet compressed explanation $S$ through optimization problem
\begin{equation} \label{eq:USIB}
    \max_{S}\mathcal{L}_{USIB}=I(Z;S)-\beta I(G;S).
\end{equation}
\end{definition}
By optimizing the USIB objective, one can make a trade-off between the informativeness and compression of explanatory subgraphs. However, it is notoriously intractable to optimize the USIB objective because the mutual information involves integral on high-dimensional data, \ie, $Z$, $S$, and $G$. 
As a result, a mutual information estimation method is necessary to be exploited. 

\subsection{Optimization for USIB}
We tackle the two terms $I(Z;S)$ and $I(G;S)$ in the objective of USIB separately. 
\paragraph{Maximizing $\bm{I(Z;S)}$.}
We adopt Jensen-Shannon MI estimator~\cite{hjelm2018learning,nowozin2016f} to assign an approximate lower bound for $I(Z;S)$, \ie,
	\begin{equation}\label{Ihat}
	     \hat{I}^{JSD}(Z;S) := \sup_{f_\phi}
	    \mathbb{E}_{p(S,Z)} \left[-sp\left(-f_{\phi}\left(S,Z\right)\right)\right] -\mathbb{E}_{p(S),p(Z)} \left[sp\left(f_{\phi}\left(S,Z\right)\right)\right]
	\end{equation}
where $sp(x)=\log(1+e^x)$ is the softplus function; Function $f_{\phi}:\mathcal{S} \times \mathcal{Z} \rightarrow \mathbb{R}$ with parameters $\phi$
is learned to discriminate whether an instance of $S$ and an instance of $Z$ are sampled from the joint distribution or not. It is implemented with function composite of $\text{MLP}_{\phi_1}$ and $\text{GNN}_{\phi_2}$, \ie,
\begin{equation}
    f_{\phi}\left(S^{(k)},Z^{(k)}\right) = \text{MLP}_{\phi_1}\left(\text{GNN}_{\phi_2}\left(S^{(k)}\right)||Z^{(k)}\right)
\end{equation}
where $\phi=\{\phi_1,\phi_2\}$; $||$ refers to the concatenation operator.
Note that the prior distributions $p(S,Z)$ and $p(Z)$ are usually unreachable in practice. In conjunction with Monte Carlo sampling to approximate the prior distributions, we reach an approximate lower bound of \cref{Ihat} by
\begin{equation} \label{eq:l1}
	\begin{aligned}	
       \max_\phi \mathcal{L}_{1}(\phi,S) = \frac{1}{K}\sum_{k=1}^K-sp\left(-f_{\phi}\left(S^{(k)},Z^{(k)}\right)\right) 
	    - \frac{1}{K}\sum_{k=1,m\neq k}^K sp\left(f_{\phi}\left(S^{(k)},Z^{(m)}\right)\right)
	\end{aligned}
\end{equation}
where $K$ is the number of samples. $(S^{(k)}, Z^{(k)})$ is sampled form joint distribution $p(S,Z)$ while $(S^{(k)},Z^{(m)})$ is independently sampled from the marginal distributions $p(S)$ and $p(Z)$, respectively \footnote{In practice, we sample $(S^{(k)}, Z^{(m)})$ by randomly permutating $(S^{(k)}, Z^{(k)})$ pairs sampled from the joint distribution.}.

\paragraph{Minimizing $\bm{I(G;S)}$.} Note that the entropy of explanatory subgraph $H(S)= \mathbb{E}_{p(S)}[-\log p(S)]$ provides an upper bound for $I(G;S)$ since the inequality $I(G;S) = H(S) - H(S|G) \leq H(S)$ holds. 
However, it is intractable to calculate the entropy because the prior distribution of $S$ is unknown in practice. 
To address this issue, we consider a relaxation and assume that the explanatory graph is a Gilbert random graph~\cite{gilbert1959random} where edges are conditionally independent to each other. Specifically, let $(i,j) \in \mathcal{E}$ denote the edge of graph $G$, and $e_{i,j}\sim\text{Bernoulli}(\mu_{i,j})$ be a binary variable indicating whether the edge $(i,j)$ is selected for subgraph $S$. 
Thus, the probability of subgraph $S$ is factorized as $p(S)=\prod_{(i,j)\in \mathcal{E}}p(e_{i,j})$, where $p(e_{i,j})=\mu_{i,j}^{e_{i,j}}(1-\mu_{i,j})^{1-e_{i,j}}$.
In this way, we can arrive at an approximate upper bound for $I(G;S)$ with Monte Carlo sampling, which is denoted by
\begin{equation}\label{eq:l2}
     \mathcal{L}_2(S) = -\frac{1}{K}\sum_{k=1}^K\sum_{(i,j)\in \mathcal{E}}e_{i,j}^{(k)}\log \mu_{i,j}^{(k)} + (1-e_{i,j}^{(k)})\log (1-\mu_{i,j}^{(k)})
\end{equation}

\begin{figure}[t] 
\centering
\includegraphics[width=0.99\textwidth]{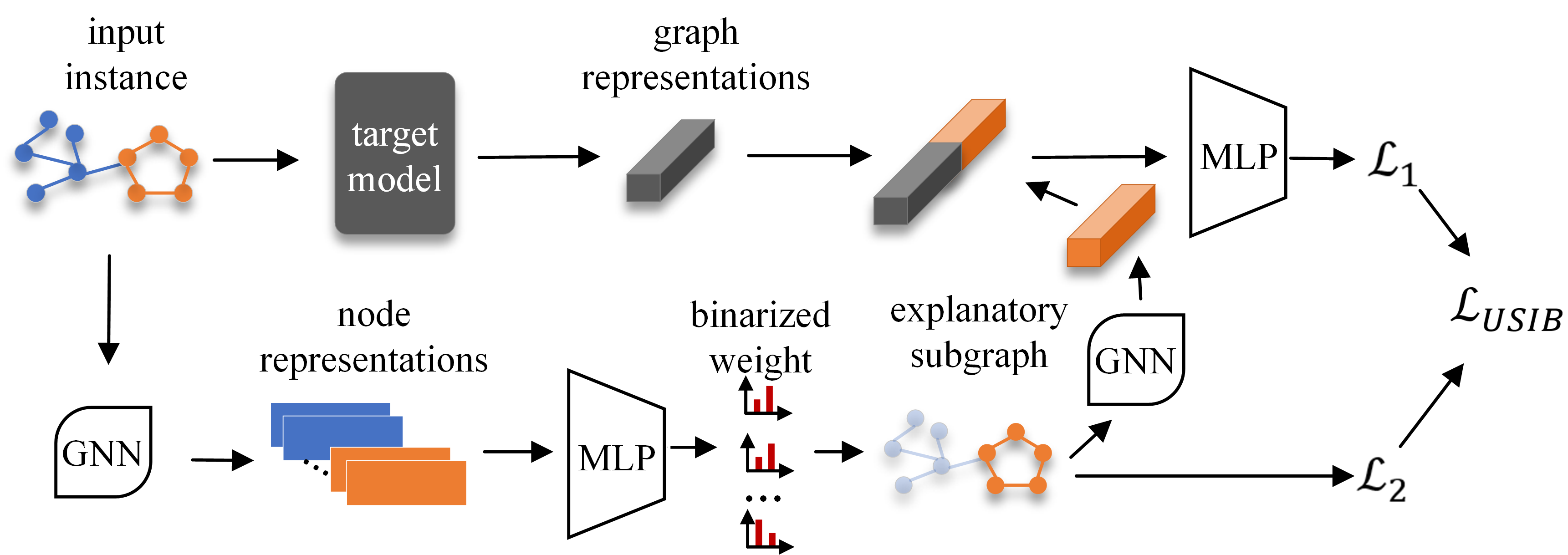} 
\caption{An illustration of our proposed USIB.} 
\vspace{-0.3cm}
\label{fig:model}
\end{figure}

\paragraph{The reparameterization trick.}
Gradients-based optimization methods may fail to optimize \cref{eq:l1} and \cref{eq:l2} because of the non-differentiable sampling process and the discrete nature of subgraph structure. On this account, we follow the Gumbel-Softmax reparametrization trick~\cite{maddison2017concrete,DBLP:conf/iclr/JangGP17} and relax the binary variables $e_{i,j}$ to a continuous edge weight variables $\hat{e}_{i,j}=\sigma((\log \epsilon - \log(1-\epsilon) + w_{i,j})/\tau) \in[0,1]$, where $\sigma(\cdot)$ is the sigmoid function; $\epsilon \sim \text{Uniform}(0,1)$; $\tau$ is the temperature hyper-parameter such that 
$\lim_{\tau\to 0}p(\hat{e}_{i,j}=1)=\sigma(w_{i,j})$; $w_{i,j}$ is the latent variables which is calculated by a neural network following previous work~\cite{luo2020parameterized},
\begin{equation}\label{eq:cal_e}
	 w_{i,j}^{(k)} = \text{MLP}_{\theta_1}\left(\mathbf{z}_i^{(k)} ||\mathbf{z}_j^{(k)}\right)\ \text{with}\ \mathbf{z}_i^{(k)} = \text{GNN}_{\theta_2}\left(G^{(k)}, i\right),i=1,2,\cdots
\end{equation}
where $\mathbf{z}_i^{(k)}$ denotes the node representations of node $i$.  
For better representation, we denote $\theta = \{\theta_1, \theta_2\}$, and generate the relaxed subgraph $\hat{S}$ by 
$\hat{S}^{(k)} = g_\theta(G^{(k)})$\footnote{$g_\theta$ computes $w_{i,j}$ first and then generates the subgraph $\hat{S}^{(k)}$ by combining the relaxed edge weights and $G^{(k)}$ into a weighted graph.}. Let $\mu^{(k)}_{i,j}=\sigma(w^{(k)}_{i,j})$, the objective in \cref{eq:l2} can be rewritten as
\begin{equation}
    \mathcal{L}_2\left(g_\theta\left(G^{(k)}\right)\right) = -\frac{1}{K}\sum_{k=1}^K\sum_{(i,j)\in \mathcal{E}}\hat{e}_{i,j}^{(k)}\log \sigma\left(w_{i,j}^{(k)}\right) + \left(1-\hat{e}_{i,j}^{(k)}\right)\log \left(1- \sigma\left(w_{i,j}^{(k)}\right)\right).
\end{equation}
In summary, we rewrite the USIB optimization problem \cref{eq:USIB} as
	\begin{equation}\label{eq:final_loss}
		\max_{\phi,\theta} \mathcal{L}_{USIB}\left(\phi, \theta, G\right) = \mathcal{L}_1\left(\phi, g_\theta\left(G^{(k)}\right)\right) - \beta * \mathcal{L}_2\left(g_\theta\left(G^{(k)}\right)\right).
	\end{equation} 
An overview of our method is shown in \cref{fig:model}. We first generate the explanatory subgraphs by a neural network, then another network is involved to estimate the mutual information between explanatory subgraphs and graph representations. Finally, the subgraph generator and the mutual information estimator are optimized collaboratively. The final explanatory subgraphs can be achieved by selecting edges with the top-$n$ edge weights (\ie, $\hat{e}_{i,j}^{(k)}$). Detailed algorithms can be found in the Appendix.

\section{Theoretical analysis}\label{sec:theoretical}
In this section, we theoretically analyze the connection between representations and explanatory subgraphs on the label space.
Before analyzing the connection, we first introduce two definitions to evaluate the expressiveness and robustness of representations.
\begin{definition}
    \textbf{Sufficiency}: Considering a specific downstream task with label $Y$, the representation $Z$ of graph $G$ is sufficient for $Y$ if and only if $I(G;Y|Z)=0$.
\end{definition}
A sufficient representation $Z$ indicates that it is capable of predicting $Y$ at least as accurately as the original graph $G$. Based on this definition, \cref{th:sufficient} holds (see proof in Appendix B).
\begin{prop}
    \label{th:sufficient}
        Let $G$ and $Y$ be random variables with joint distribution $p(G,Y)$, and $Z$ be a representation of $G$. Considering a Markov Chain $Y\rightarrow G\rightarrow Z$, we assume that $Z$ is conditionally independent from $Y$ once $G$ is given. Then, $Z$ is sufficient for $Y$ if and only if $I(Z;Y)=I(G;Y)$.
\end{prop}
The \textit{sufficiency} of representations evaluates the expressiveness of unsupervised representations, \ie, for graph representations that are sufficient for a specific task, no information regarding the task is lost in the embedding procedure. Thus the unsupervised representations are expressive enough for the task.
\begin{definition}
    \textbf{Necessity}: Considering a specific downstream task with label $Y$, the representation $Z$ of graph $G$ is necessary for $Y$ if and only if $I(G;Z|Y)=0$.
\end{definition}
A necessary representation indicates that $Z$ only extracts information relevant to $Y$ from the graph $G$. According to this definition, the following \cref{th:necessary} holds (see proof in Appendix B).
\begin{prop}\label{th:necessary}
     Let $G$ and $Y$ be random variables with joint distribution $p(G,Y)$, and $Z$ be a representation of $G$. Considering a Markov Chain: $(G_N,Y) \rightarrow G \rightarrow Z$ where $G_N$ collects all label-irrelevant information in $G$, $Z$ is necessary for $Y$ if and only if $I(G_N;Z) = 0$.
\end{prop}
The \textit{necessity} of representations evaluates the robustness of unsupervised representations, \ie, graph representations are unacted on label-irrelevant information, and thus the representations can be robust against noise information.

Given the definition of \textit{sufficiency} and \textit{necessity}, we have the following \cref{th:last} that provides a theoretical connection between the graph representations and explanatory subgraphs on the label space (see proof in Appendix B).
\begin{theorem}\label{th:last}
	Let $G$ and $Y$ be random variables with joint distribution $p(G,Y)$. Given unsupervised graph representation $Z$ and its explanations $S$ in graph $G$, we assume that $S$ is conditionally independent from $Y$ and $Z$ when $G$ is observed. We have,
	\begin{enumerate}
	\item if $Z$ is sufficient for $Y$, then $I(S;Z) \geq I(S;Y)$;
	\item if $Z$ is necessary for $Y$, then $I(S;Z) \leq I(S;Y)$;
	\item if $Z$ is sufficient and necessary for $Y$, then $I(S;Z) = I(S;Y)$;
	\end{enumerate}
\end{theorem}
\cref{th:last} provides theoretical support for our USIB explanation method, \ie, for representations that are expressive and robust enough, one can explore explanatory subgraphs that are highly relevant to the ground-truth labels by maximizing the mutual information between representations and explanatory subgraphs. In contrast, if representations are uninformative and fragile, the fidelity of explanatory subgraphs is not guaranteed. We experimentally analyze the influence of representations' expressiveness and robustness in the next section.


\section{Experiments} \label{sec:exp}
In this section,  we empirically evaluate the effectiveness and superiority of our proposed method by answering the following questions.
\begin{itemize}[leftmargin=*]
	    \item \textbf{RQ1} How does our proposed method perform compared to other baseline explainers?
	    \item \textbf{RQ2} Does expressiveness and robustness of representations affect the fidelity of explanatory subgraphs in agreement with the theoretical analysis?
\end{itemize}
\subsection{Experimental settings}

\paragraph{Datasets.} We take four benchmark datasets for experiments, including Mutagenicity, NCI1, PROTEINS~\cite{Morris+2020}, and BA3~\cite{wx2021refine}. The first two are molecule graph datasets, while PROTEINS is a bioinformatics graph dataset. BA3 is a synthetic dataset created by Wang \etal~\cite{wx2021refine}. Specifically, the Barabasi-Albert (BA) graphs are adopted as the base and attach each base with one of three motifs: house, cycle, and grid. 
The statistical details for these datasets can be found in Appendix D.

\paragraph{Target models.} We consider three state-of-the-art target models to generate graph representations for explanation: Infograph~\cite{You2020GraphCL}, GCL~\cite{You2020GraphCL}, and ADGCL~\cite{suresh2021adversarial}. Note that all these target models adopt the same GNN encoder architecture for a fair comparison. The details of implementation can be found in Appendix D.

\paragraph{Baselines.} To our best knowledge, there is no previous work focusing on the explanation for unsupervised graph representation learning.
We train an MLP classifier based on representations generated by the target models and then compose the target models and the classifiers as end-to-end graph classifiers. Following this strategy, previous explanation methods for supervised setting, such as GNNExplainer~\cite{ying2019gnnexplainer}, PGExplainer~\cite{luo2020parameterized}, PGM-Explainer~\cite{vu2020pgm}, and Refine~\cite{wx2021refine}, could be adopted as baselines. Moreover, to provide a fair comparison without the involvement of label information, we also adapt several typical explanation methods for CNNs to our unsupervised graph representation explanation setting. Specifically, we adopt three methods which involve gradient-like scores by conducting backpropagation from the representations' $l_2$ norm to the input, such as IG~\cite{sundararajan2017axiomatic}, GradCam~\cite{pope2019explainability}, and SA~\cite{baldassarre2019explainability}. 

\paragraph{Evaluation metrics.}
It is challenging to evaluate explanatory subgraphs quantitatively because the ground-truth explanations in practice are usually unreachable. Following previous works~\cite{ying2019gnnexplainer,wx2021refine,chen2018learning,liang2020adversarial}, we adopt the following two metrics: (1) \textit{Predictive Accuracy (ACC@$r$)}.  For the explanatory subgraphs, which are crucial for the target GNN encoders, the corresponding subgraph representations should behave similarly to representations of the original graph on downstream tasks, \eg, graph classification. 
To evaluate the fidelity of explanatory subgraphs, we feed the subgraphs to the target models first and get the subgraph representations. Then we train logistic classifiers on the subgraph representations and report the 10-fold cross-validation accuracy as ACC@$r$ for explanatory subgraphs with edge selection ratio $r$. Moreover we further denote ACC-AUC as the area under the ACC curve over different selection ratios $r\in \{0.1, 0.2, \cdots,  0.9\}$. (2) \textit{Recall@$n$}. Ground-truth explanations for the synthetic dataset BA3 are reachable. Thus we adopt the recall metric to evaluate explanatory subgraphs. It is formulated as $\text{Recall@}n=\mathbb{E}_{\mathcal{S}}\left[|S\bigcap S^*|/|S^*|\right]$, where $n$ is the number of edges involved in the explanatory subgraphs; $S^*$ denotes the ground-truth explanations. Note that each experiment is repeated $10$ times to reduce randomness. 

\subsection{Quantitative evaluations}

\begin{figure}[t] 
\centering 
\setlength{\abovecaptionskip}{0.cm}
\includegraphics[width=0.99\textwidth]{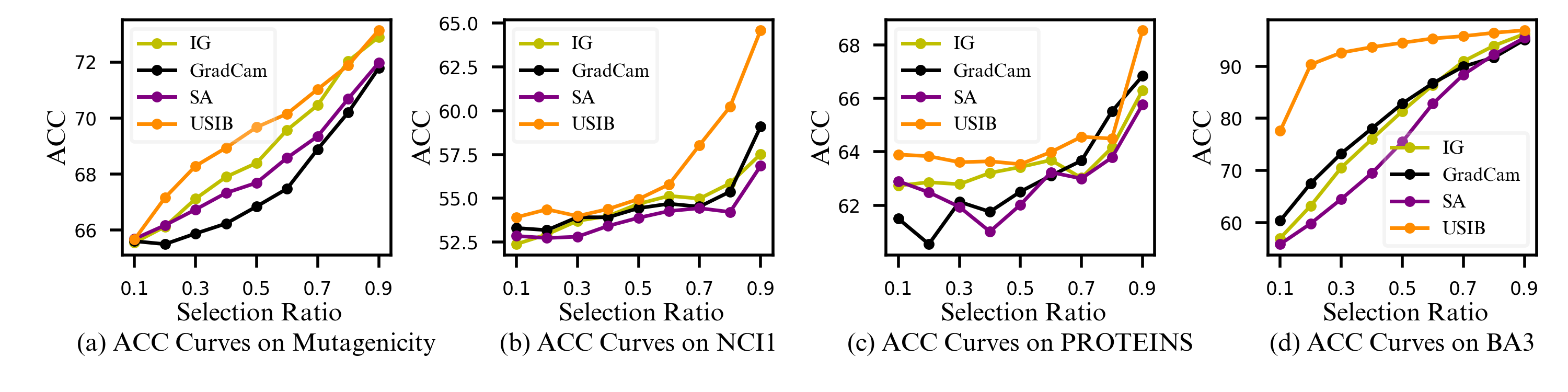} 
\caption{ACC curve of different explainers on Infograph.}
\label{fig:infograph_acc_curve} 
\end{figure}
\begin{table*}[t]
	\centering
	\normalsize
	\setlength{\tabcolsep}{5pt}
	\caption{Comparison of our USIB and other baseline explainers. The relative improvement is calculated on unsupervised setting methods. \textbf{Bold} indicates the best results.}\label{tab:acc_auc}
	\resizebox{0.99\textwidth}{!}{
		\begin{tabular}{lllccccc}
			\toprule
			\multirow{2}{*}{Target Model}&\multirow{2}{*}{Setting}&\multirow{2}{*}{Method}&
			Mutagenicity&NCI1&PROTEINS&\multicolumn{2}{c}{BA3}\cr
			\cmidrule(lr){7-8}
			&&&ACC-AUC&ACC-AUC&ACC-AUC&ACC-AUC&Recall@5\cr
			\midrule
			\multirow{9}{*}{Infograph}&\multirow{4}{*}{Supervised}
			&PG-Eplainer&$69.29\pm 2.10$&$\bf{59.93\pm 1.42}$&$65.50\pm 2.32$&$89.99\pm 3.81$&$26.49\pm 3.27$\cr
            &&GNNExplainer&$68.99\pm 2.08$&$54.68\pm 1.34$&$62.16\pm 2.20$&$74.78\pm 3.94$&$18.80\pm 0.92$\cr
            &&PGM-Explainer&$66.93\pm 0.42$&$54.80\pm 1.26$&$64.64\pm 1.97$&$83.70\pm 1.85$&$24.76\pm 0.07$\cr
            &&ReFine&$67.45\pm 0.97$&$57.88\pm 1.30$&$\bf{66.28\pm 1.66}$&$84.78\pm 6.59$&$17.88\pm 8.79$\cr
            \cmidrule(lr){2-8}&\multirow{5}{*}{Unsupervised}
            &IG&$68.87\pm 1.52$&$54.65\pm 0.94$&$63.30\pm 2.25$&$79.52\pm 2.18$&$16.22\pm 5.11$\cr
            &&GradCam&$67.60\pm 0.93$&$54.79\pm 0.71$&$63.12\pm 2.36$&$80.62\pm 3.84$&$18.74\pm 1.74$\cr
            &&SA&$68.24\pm 0.97$&$53.94\pm 0.65$&$62.71\pm 2.16$&$76.02\pm 3.07$&$16.35\pm 3.01$\cr
            &&USIB&$\bf{69.55\pm 0.96}$&$56.70\pm 1.65$&$64.46\pm 2.75$&$\bf{92.55\pm 3.28}$&$\bf{27.16\pm 4.46}$\cr
            \cmidrule(lr){3-8}& &Relative Improvement
            & 0.99\% & 3.49\% & 1.83\%& 14.80\%& 44.93\%\cr
			\bottomrule
		\end{tabular}
	}
\end{table*}

\paragraph{Effectiveness of USIB.} To investigate the effectiveness of our proposed USIB explanation method, we present the ACC curve over different selection ratios in \cref{fig:infograph_acc_curve}, and report the results on ACC-AUC and Recall@$5$ in \cref{tab:acc_auc}. We only present the results on Infograph because of page limitation and present the results on other target models in Appendix D. The experimental results show that our proposed USIB achieves superior performance in the unsupervised setting and is comparable to baseline methods in the supervised setting. 
Specifically, USIB achieves the best performance on Mutagenicity and BA3 among all baseline methods and outperforms other unsupervised methods on NCI1 and PROTEINS. However, there is still a gap between our USIB and supervised baseline methods on NCI1 and PROTEINS because of the involvement of label information. It is noteworthy that a significant relative improvement is achieved (\ie, $14.8\%$ on ACC-AUC and $10.44\%$ on Recall@$5$) on the synthetic dataset BA3. It indicates that the performance of USIB is related to the expressiveness of target models (three target models achieve more than $90\%$ ACC on BA3, which is much better than other datasets). 

\begin{figure}
    \centering
    \setlength{\abovecaptionskip}{0.cm}
    \includegraphics[width=0.8\textwidth]{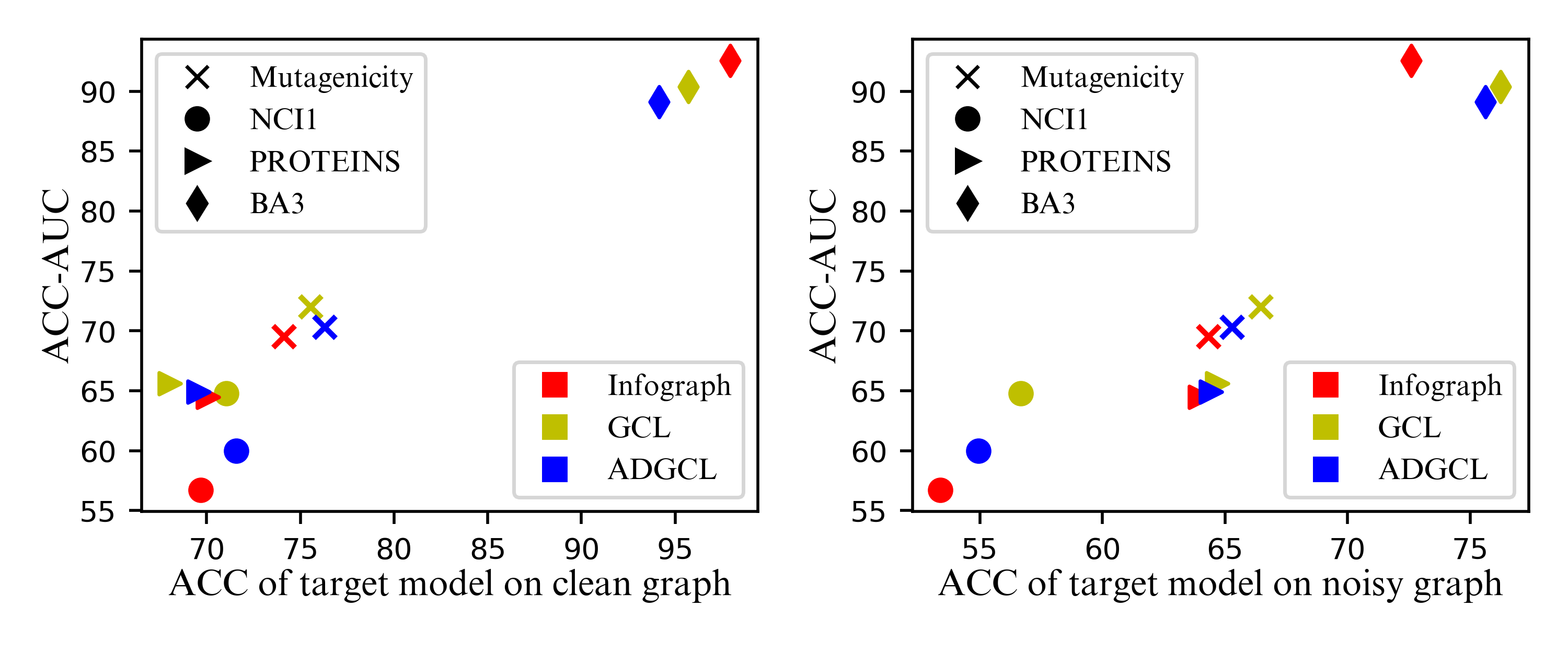}
    \caption{Influence of target models' expressiveness and robustness. We distinguish different target models and datasets with different colors and markers separately.}\label{fig:rob_acc}
\end{figure}

\paragraph{Influence of representations' expressiveness and robustness.} 
We quantitatively evaluate the expressiveness and robustness by the downstream graph classification accuracy (\ie, ACC) on representations extracted respectively from the clean graphs and noisy graphs. The noisy graphs are generated by randomly adding $|\mathcal{E}|$ noise edges to the clean graphs.  
To investigate how representations' expressiveness and robustness affect the fidelity of explanatory subgraphs, we show a scatter diagram in \cref{fig:rob_acc}, where each point denotes a target model 
on a specific dataset. 
From \cref{fig:rob_acc}, we can observe that: (1) There is a positive correlation between representations' expressiveness/robustness and explanatory subgraphs' fidelity. Representations achieving higher ACC on clean and noisy graphs usually lead to explanatory subgraphs with higher ACC-AUC. This observation is consistent with our theoretical analysis. (2) Expressiveness matters more than robustness when there is a significant gap on expressiveness. For example, we can observe that the fidelity of explanatory subgraphs on BA3 is positively related to the expressiveness while robustness makes less influence. (3) Robustness matters when there is only a slight gap on expressiveness. Specifically, the fidelity of explanatory subgraphs on Mutagenicity, NCI1, and PROTEINS is positively related to representations' robustness since there is only a slight gap of expressiveness over clean graphs.

\subsection{Qualitative analysis}
\begin{figure}
    \centering
    \includegraphics[width=0.95\textwidth]{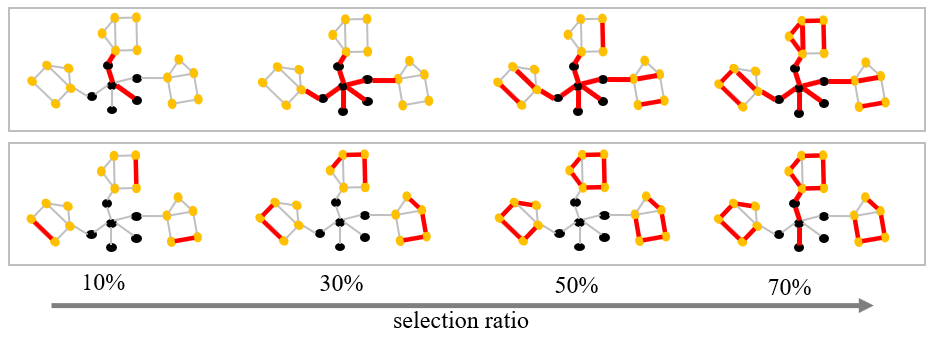}
    \caption{Qualitative results in BA3. The first row shows the explanation results of USIB on untrained Infograph, while the second row is the explanation results on fine-tuned Infograph. The yellow nodes denote the house motifs and the black nodes denote the random base graph. Explanation edges are colored as red.}\label{fig:qualitative}
    \vspace{-0.5cm}
\end{figure}
We present the qualitative results on the synthetic dataset BA3 in \cref{fig:qualitative}. Specifically, we compare the explanation results of our USIB on untrained Infograph and fine-tuned Infograph qualitatively. We can clearly observe that the untrained Infograph focuses more on the random base graph. This can be attributed that the untrained Infograph recursively incorporates node's information with random weights, and thus pays more attention to the central nodes in the graph, \ie, the base graph. In contrast, the second row shows that the fine-tuned Infograph transfers its attention to the house motifs after the unsupervised training, which reveals why representations perform well on the downstream tasks. Moreover, we surprisingly find that Infograph's performance does not rely on the completeness of motifs, \eg, Infograph performs well even with only 30\% edges according to \cref{fig:infograph_acc_curve}. A rational explanation is the redundancy of graph information, \ie, node features may preserve redundant information that is also contained in structural features. The redundancy may jeopardize the evaluation of explanation methods~\cite{faber2021comparing}. We consider it a future direction to construct datasets carefully designed for unsupervised graph representation explanation. We can also observe a limitation of USIB that it considers edges independently but ignores the substructures of graphs whose importance is emphasized in previous work~\cite{yuan2021explainability}.

\section{Conclusion and future work}
We study an unexplored explanation problem: the explanation for unsupervised graph representation learning. We advance the IB principle to tackle the explanation problem, which leads to a novel explanation method USIB. Moreover, we theoretically analyze the connection between representations and explanatory subgraphs on the label space, which reveals that expressiveness and robustness benefit the fidelity of explanatory subgraphs. Extensive results on four datasets and three target models demonstrate the superiority of our methods and the validity of our theoretical analysis. As a future direction, we consider the counterfactual explanation~\cite{lucic2021cf} for unsupervised representation learning and explore whether there is a connection between explanation and adversarial examples~\cite{zugner2018adversarial,dai2018adversarial,wang2020scalable}.

\bibliographystyle{IEEEtran}
\bibliography{main}

\begin{thebibliography}{10}
\providecommand{\url}[1]{#1}
\csname url@samestyle\endcsname
\providecommand{\newblock}{\relax}
\providecommand{\bibinfo}[2]{#2}
\providecommand{\BIBentrySTDinterwordspacing}{\spaceskip=0pt\relax}
\providecommand{\BIBentryALTinterwordstretchfactor}{4}
\providecommand{\BIBentryALTinterwordspacing}{\spaceskip=\fontdimen2\font plus
\BIBentryALTinterwordstretchfactor\fontdimen3\font minus
  \fontdimen4\font\relax}
\providecommand{\BIBforeignlanguage}[2]{{%
\expandafter\ifx\csname l@#1\endcsname\relax
\typeout{** WARNING: IEEEtran.bst: No hyphenation pattern has been}%
\typeout{** loaded for the language `#1'. Using the pattern for}%
\typeout{** the default language instead.}%
\else
\language=\csname l@#1\endcsname
\fi
#2}}
\providecommand{\BIBdecl}{\relax}
\BIBdecl

\bibitem{kipf2017semi}
T.~N. Kipf and M.~Welling, ``Semi-supervised classification with graph
  convolutional networks,'' in \emph{International Conference on Learning
  Representations (ICLR)}, 2017.

\bibitem{velickovic2018graph}
\BIBentryALTinterwordspacing
P.~Veli{\v{c}}kovi{\'{c}}, G.~Cucurull, A.~Casanova, A.~Romero, P.~Li{\`{o}},
  and Y.~Bengio, ``{Graph Attention Networks},'' \emph{International Conference
  on Learning Representations}, 2018, accepted as poster. [Online]. Available:
  \url{https://openreview.net/forum?id=rJXMpikCZ}
\BIBentrySTDinterwordspacing

\bibitem{liu2020towards}
M.~Liu, H.~Gao, and S.~Ji, ``Towards deeper graph neural networks,'' in
  \emph{Proceedings of the 26th ACM SIGKDD international conference on
  knowledge discovery \& data mining}, 2020, pp. 338--348.

\bibitem{hamilton2017inductive}
W.~Hamilton, Z.~Ying, and J.~Leskovec, ``Inductive representation learning on
  large graphs,'' \emph{Advances in neural information processing systems},
  vol.~30, 2017.

\bibitem{ying2018hierarchical}
Z.~Ying, J.~You, C.~Morris, X.~Ren, W.~Hamilton, and J.~Leskovec,
  ``Hierarchical graph representation learning with differentiable pooling,''
  \emph{Advances in neural information processing systems}, vol.~31, 2018.

\bibitem{xu2018how}
\BIBentryALTinterwordspacing
K.~Xu, W.~Hu, J.~Leskovec, and S.~Jegelka, ``How powerful are graph neural
  networks?'' in \emph{International Conference on Learning Representations},
  2019. [Online]. Available: \url{https://openreview.net/forum?id=ryGs6iA5Km}
\BIBentrySTDinterwordspacing

\bibitem{zhang2018link}
M.~Zhang and Y.~Chen, ``Link prediction based on graph neural networks,''
  \emph{Advances in neural information processing systems}, vol.~31, 2018.

\bibitem{cai2021line}
L.~Cai, J.~Li, J.~Wang, and S.~Ji, ``Line graph neural networks for link
  prediction,'' \emph{IEEE Transactions on Pattern Analysis and Machine
  Intelligence}, 2021.

\bibitem{ying2019gnnexplainer}
Z.~Ying, D.~Bourgeois, J.~You, M.~Zitnik, and J.~Leskovec, ``Gnnexplainer:
  Generating explanations for graph neural networks,'' \emph{Advances in neural
  information processing systems}, vol.~32, 2019.

\bibitem{luo2020parameterized}
D.~Luo, W.~Cheng, D.~Xu, W.~Yu, B.~Zong, H.~Chen, and X.~Zhang, ``Parameterized
  explainer for graph neural network,'' \emph{Advances in neural information
  processing systems}, vol.~33, pp. 19\,620--19\,631, 2020.

\bibitem{doshi2017towards}
F.~Doshi-Velez and B.~Kim, ``Towards a rigorous science of interpretable
  machine learning,'' \emph{arXiv preprint arXiv:1702.08608}, 2017.

\bibitem{pope2019explainability}
P.~E. Pope, S.~Kolouri, M.~Rostami, C.~E. Martin, and H.~Hoffmann,
  ``Explainability methods for graph convolutional neural networks,'' in
  \emph{Proceedings of the IEEE/CVF Conference on Computer Vision and Pattern
  Recognition}, 2019, pp. 10\,772--10\,781.

\bibitem{yuan2020explainability}
H.~Yuan, H.~Yu, S.~Gui, and S.~Ji, ``Explainability in graph neural networks: A
  taxonomic survey,'' \emph{arXiv preprint arXiv:2012.15445}, 2020.

\bibitem{vu2020pgm}
M.~Vu and M.~T. Thai, ``Pgm-explainer: Probabilistic graphical model
  explanations for graph neural networks,'' \emph{Advances in neural
  information processing systems}, vol.~33, pp. 12\,225--12\,235, 2020.

\bibitem{perozzi2014deepwalk}
B.~Perozzi, R.~Al-Rfou, and S.~Skiena, ``Deepwalk: Online learning of social
  representations,'' in \emph{Proceedings of the 20th ACM SIGKDD international
  conference on Knowledge discovery and data mining}, 2014, pp. 701--710.

\bibitem{tang2015line}
J.~Tang, M.~Qu, M.~Wang, M.~Zhang, J.~Yan, and Q.~Mei, ``Line: Large-scale
  information network embedding,'' in \emph{Proceedings of the 24th
  international conference on world wide web}, 2015, pp. 1067--1077.

\bibitem{grover2016node2vec}
A.~Grover and J.~Leskovec, ``node2vec: Scalable feature learning for
  networks,'' in \emph{Proceedings of the 22nd ACM SIGKDD international
  conference on Knowledge discovery and data mining}, 2016, pp. 855--864.

\bibitem{kipf2016variational}
T.~N. Kipf and M.~Welling, ``Variational graph auto-encoders,'' \emph{NIPS
  Workshop on Bayesian Deep Learning}, 2016.

\bibitem{yu2021recognizing}
J.~Yu, T.~Xu, Y.~Rong, Y.~Bian, J.~Huang, and R.~He, ``Recognizing predictive
  substructures with subgraph information bottleneck,'' \emph{IEEE Transactions
  on Pattern Analysis and Machine Intelligence}, 2021.

\bibitem{wu2020graph}
T.~Wu, H.~Ren, P.~Li, and J.~Leskovec, ``Graph information bottleneck,''
  \emph{Advances in Neural Information Processing Systems}, vol.~33, pp.
  20\,437--20\,448, 2020.

\bibitem{baldassarre2019explainability}
F.~Baldassarre and H.~Azizpour, ``Explainability techniques for graph
  convolutional networks,'' in \emph{International Conference on Machine
  Learning (ICML) Workshops, 2019 Workshop on Learning and Reasoning with
  Graph-Structured Representations}, 2019.

\bibitem{schnake2020higher}
T.~Schnake, O.~Eberle, J.~Lederer, S.~Nakajima, K.~T. Schutt, K.-R. Mueller,
  and G.~Montavon, ``Higher-order explanations of graph neural networks via
  relevant walks,'' \emph{IEEE Transactions on Pattern Analysis and Machine
  Intelligence}, 2021.

\bibitem{sundararajan2017axiomatic}
M.~Sundararajan, A.~Taly, and Q.~Yan, ``Axiomatic attribution for deep
  networks,'' in \emph{International conference on machine learning}.\hskip 1em
  plus 0.5em minus 0.4em\relax PMLR, 2017, pp. 3319--3328.

\bibitem{schwab2019cxplain}
P.~Schwab and W.~Karlen, ``Cxplain: Causal explanations for model
  interpretation under uncertainty,'' \emph{Advances in Neural Information
  Processing Systems}, vol.~32, 2019.

\bibitem{wx2021refine}
X.~Wang, Y.-X. Wu, A.~Zhang, X.~He, and T.-S. Chua, ``Towards multi-grained
  explainability for graph neural networks,'' in \emph{Proceedings of the 35th
  Conference on Neural Information Processing Systems}, 2021.

\bibitem{yuan2020xgnn}
H.~Yuan, J.~Tang, X.~Hu, and S.~Ji, ``Xgnn: Towards model-level explanations of
  graph neural networks,'' in \emph{Proceedings of the 26th ACM SIGKDD
  International Conference on Knowledge Discovery \& Data Mining}, 2020, pp.
  430--438.

\bibitem{sun2019infograph}
F.-Y. Sun, J.~Hoffman, V.~Verma, and J.~Tang, ``Infograph: Unsupervised and
  semi-supervised graph-level representation learning via mutual information
  maximization,'' in \emph{International Conference on Learning
  Representations}, 2019.

\bibitem{You2020GraphCL}
\BIBentryALTinterwordspacing
Y.~You, T.~Chen, Y.~Sui, T.~Chen, Z.~Wang, and Y.~Shen, ``Graph contrastive
  learning with augmentations,'' in \emph{Advances in Neural Information
  Processing Systems}, H.~Larochelle, M.~Ranzato, R.~Hadsell, M.~F. Balcan, and
  H.~Lin, Eds., vol.~33.\hskip 1em plus 0.5em minus 0.4em\relax Curran
  Associates, Inc., 2020, pp. 5812--5823. [Online]. Available:
  \url{https://proceedings.neurips.cc/paper/2020/file/3fe230348e9a12c13120749e3f9fa4cd-Paper.pdf}
\BIBentrySTDinterwordspacing

\bibitem{suresh2021adversarial}
S.~Suresh, P.~Li, C.~Hao, and J.~Neville, ``Adversarial graph augmentation to
  improve graph contrastive learning,'' \emph{NeurIPS}, 2021.

\bibitem{tishby1999information}
N.~TISHBY, ``The information bottleneck method,'' in \emph{Proc. 37th Annual
  Allerton Conference on Communications, Control and Computing, 1999}, 1999,
  pp. 368--377.

\bibitem{tishby2015deep}
N.~Tishby and N.~Zaslavsky, ``Deep learning and the information bottleneck
  principle,'' in \emph{2015 ieee information theory workshop (itw)}.\hskip 1em
  plus 0.5em minus 0.4em\relax IEEE, 2015, pp. 1--5.

\bibitem{hjelm2018learning}
R.~D. Hjelm, A.~Fedorov, S.~Lavoie-Marchildon, K.~Grewal, P.~Bachman,
  A.~Trischler, and Y.~Bengio, ``Learning deep representations by mutual
  information estimation and maximization,'' in \emph{International Conference
  on Learning Representations}, 2018.

\bibitem{nowozin2016f}
S.~Nowozin, B.~Cseke, and R.~Tomioka, ``f-gan: Training generative neural
  samplers using variational divergence minimization,'' \emph{Advances in
  neural information processing systems}, vol.~29, 2016.

\bibitem{gilbert1959random}
E.~N. Gilbert, ``Random graphs,'' \emph{The Annals of Mathematical Statistics},
  vol.~30, no.~4, pp. 1141--1144, 1959.

\bibitem{maddison2017concrete}
C.~Maddison, A.~Mnih, and Y.~Teh, ``The concrete distribution: A continuous
  relaxation of discrete random variables,'' in \emph{Proceedings of the
  international conference on learning Representations}.\hskip 1em plus 0.5em
  minus 0.4em\relax International Conference on Learning Representations, 2017.

\bibitem{DBLP:conf/iclr/JangGP17}
\BIBentryALTinterwordspacing
E.~Jang, S.~Gu, and B.~Poole, ``Categorical reparameterization with
  gumbel-softmax,'' in \emph{5th International Conference on Learning
  Representations, {ICLR} 2017, Toulon, France, April 24-26, 2017, Conference
  Track Proceedings}.\hskip 1em plus 0.5em minus 0.4em\relax OpenReview.net,
  2017. [Online]. Available: \url{https://openreview.net/forum?id=rkE3y85ee}
\BIBentrySTDinterwordspacing

\bibitem{Morris+2020}
\BIBentryALTinterwordspacing
C.~Morris, N.~M. Kriege, F.~Bause, K.~Kersting, P.~Mutzel, and M.~Neumann,
  ``Tudataset: A collection of benchmark datasets for learning with graphs,''
  in \emph{ICML 2020 Workshop on Graph Representation Learning and Beyond (GRL+
  2020)}, 2020. [Online]. Available: \url{www.graphlearning.io}
\BIBentrySTDinterwordspacing

\bibitem{chen2018learning}
J.~Chen, L.~Song, M.~Wainwright, and M.~Jordan, ``Learning to explain: An
  information-theoretic perspective on model interpretation,'' in
  \emph{International Conference on Machine Learning}.\hskip 1em plus 0.5em
  minus 0.4em\relax PMLR, 2018, pp. 883--892.

\bibitem{liang2020adversarial}
J.~Liang, B.~Bai, Y.~Cao, K.~Bai, and F.~Wang, ``Adversarial infidelity
  learning for model interpretation,'' in \emph{Proceedings of the 26th ACM
  SIGKDD International Conference on Knowledge Discovery \& Data Mining}, 2020,
  pp. 286--296.

\bibitem{faber2021comparing}
L.~Faber, A.~K.~Moghaddam, and R.~Wattenhofer, ``When comparing to ground truth
  is wrong: On evaluating gnn explanation methods,'' in \emph{Proceedings of
  the 27th ACM SIGKDD Conference on Knowledge Discovery \& Data Mining}, 2021,
  pp. 332--341.

\bibitem{yuan2021explainability}
H.~Yuan, H.~Yu, J.~Wang, K.~Li, and S.~Ji, ``On explainability of graph neural
  networks via subgraph explorations,'' in \emph{International Conference on
  Machine Learning}.\hskip 1em plus 0.5em minus 0.4em\relax PMLR, 2021, pp.
  12\,241--12\,252.

\bibitem{lucic2021cf}
A.~Lucic, M.~ter Hoeve, G.~Tolomei, M.~de~Rijke, and F.~Silvestri,
  ``Cf-gnnexplainer: Counterfactual explanations for graph neural networks,''
  \emph{arXiv preprint arXiv:2102.03322}, 2021.

\bibitem{zugner2018adversarial}
D.~Z{\"u}gner, A.~Akbarnejad, and S.~G{\"u}nnemann, ``Adversarial attacks on
  neural networks for graph data,'' in \emph{Proceedings of the 24th ACM SIGKDD
  International Conference on Knowledge Discovery \& Data Mining}, 2018, pp.
  2847--2856.

\bibitem{dai2018adversarial}
H.~Dai, H.~Li, T.~Tian, X.~Huang, L.~Wang, J.~Zhu, and L.~Song, ``Adversarial
  attack on graph structured data,'' in \emph{International conference on
  machine learning}.\hskip 1em plus 0.5em minus 0.4em\relax PMLR, 2018, pp.
  1115--1124.

\bibitem{wang2020scalable}
J.~Wang, M.~Luo, F.~Suya, J.~Li, Z.~Yang, and Q.~Zheng, ``Scalable attack on
  graph data by injecting vicious nodes,'' \emph{Data Mining and Knowledge
  Discovery}, vol.~34, no.~5, pp. 1363--1389, 2020.

\bibitem{federici2020learning}
M.~Federici, A.~Dutta, P.~Forr{\'e}, N.~Kushman, and Z.~Akata, ``Learning
  robust representations via multi-view information bottleneck,'' in \emph{8th
  International Conference on Learning Representations}.\hskip 1em plus 0.5em
  minus 0.4em\relax OpenReview. net, 2020.

\end{thebibliography}
\clearpage

\appendix

	\section{PROPERTIES OF MUTUAL INFORMATION}
	In this section we enumerate some properties~\cite{federici2020learning} of mutual information that are used to prove the theorems reported in this work. For any random variables $G$, $Y$, $Z$ and $S$:
	
	$(P_1)$ Positivity: 
	\begin{equation}
	    I(G;Y)\geq0, I(G;Y|Z)\geq0
	\end{equation}
	$(P_2)$ Chain rule:
	\begin{equation}
		I(G,Y;Z) = I(G;Z) + I(Y;Z|G)	
	\end{equation}
	$(P_3)$ Chain rule (Multivariate Mutual Information):
	\begin{equation}
	    I(G; Y; Z) = I(Y; Z) - I(Y; Z|G)
	\end{equation}
	\begin{equation}
	     I(G; Y; Z|S) = I(Y; Z|S) - I(Y; Z|G,S)
	\end{equation}

	\section{THEOREMS AND PROOFS}
	In the following section we prove the statements reported in the main body of the paper. We consider the following graphical model $\mathcal{G}$:
	\begin{equation}
	    \begin{aligned}
	        Y \rightarrow \ &G  \rightarrow Z \\
	        &\downarrow \\
	        & S
	    \end{aligned}
	\end{equation}
	Following Federici \etal~\cite{federici2020learning},  we state that $Z$ and $S$ are conditionally independent from any other variables in the system once $G$ is observed. As a result, whenever $Z$ is the representation of $G$ and $S$ is an explanation of $G$, we have
    \begin{equation}
        I(Z; A|G,B) = 0, \ I(S; A|G,B) = 0
    \end{equation}
     for any variable (or groups of variables) $A$ and $B$ in the system. 

	\begin{lemma} \label{appendix_lemma:up_low}
              Let $G$ be a random variable which denotes a graph, $Z$ is the representation of $G$ and $S$ denotes the regarding explanatory subgraph, we have
	       \begin{equation}
	           I(Z;S) \leq I(Z;G)
	       \end{equation} 

	       \noindent Hypothesis:
        
            $(H_1)$ $S$ is an explanation of $G$: $I(S;Z|G)=0$
            
	       
	       \noindent Thesis:
	       
	       $(T_1)$ $I(Z;S) \leq I(Z;G)$
	       
	       \begin{proof}
            \begin{equation}
            \begin{aligned}
            I(Z;S) &\overset{P_3}{=} I(Z;S;G) + I(Z;S|G)\\
            &\overset{H_1}{=} I(Z;S;G)\\
            &\overset{P_3}{=} I(Z;G) - I(Z;G|S)\\
            & \leq I(Z;G)
            \end{aligned}
            \end{equation}
        \end{proof}
	\end{lemma}

    \begin{prop}\label{prop:sufficient}

        Let $G$ and $Y$ be random variables with joint distribution $p(G,Y)$, and $Z$ be a representation of $G$. Considering a Markov Chain $Y\rightarrow G\rightarrow Z$, we assume that $Z$ is conditionally independent from $Y$ once $G$ is given. Then, $Z$ is sufficient for $Y$ if and only if $I(Z;Y)=I(G;Y)$.

        \noindent Hypothesis:
        
        $(H_1)$ $Z$ is a representation of $G$: $I(Z;Y|G)=0$
        
        \noindent Thesis:
        
        $(T_1)$ $I(G;Y|Z)=0 \iff I(Z;Y)=I(G;Y)$ 
        \begin{proof}
            \begin{equation}
            \begin{aligned}
                I(G;Y|Z) &\overset{P_3}{=} I(G;Y) - I(G;Y;Z)\\
                           &\overset{P_3}{=} I(G;Y) - I(Y;Z) + I(Y;Z|G) \\
                           &\overset{H_1}{=} I(G;Y) - I(Y;Z)
            \end{aligned}
            \end{equation}
        Thus we can have $I(G;Y|Z)=0 \iff I(Z;Y)=I(G;Y)$.
        \end{proof}
    \end{prop}
    
	\begin{prop}\label{prop:necessary}
        Let $G$ and $Y$ be random variables with joint distribution $p(G,Y)$, and $Z$ be a representation of $G$. Considering a Markov Chain: $(G_N,Y) \rightarrow G \rightarrow Z$ where $G_N$ collects all label-irrelevant information in $G$, $Z$ is necessary for $Y$ if and only if $I(G_N;Z) = 0$.

        \noindent Hypothesis:
        
        $(H_1)$ $Z$ is a representation of $G$: $I(Z;G_N|G)=0$;
        
        $(H_2)$ $G_N$ is label-irrelevant: $I(G_N;Y) = 0$;
        
        $(H_3)$ $G_N$ collects all label-irrelevant information in $G$: $I(G;Z|Y,G_N) = 0$

        \noindent Thesis:
        
        $(T_1)$ $I(Z;G|Y)=0 \iff I(Z;G_N)=0$ 
        
        \begin{proof}
            \begin{equation}
            \begin{aligned}
                I(Z;G|Y) &\overset{P_3}{=} I(Z;G) - I(Z;G;Y)\\
                		& =  I(Z;G) - I(Y,G_N;G;Z) + I(Y,G_N;G;Z) - I(Z;G;Y)\\
                		&\overset{P_3}{=} I(Z;G|Y,G_N) +  I(Y,G_N;G;Z) - I(Z;G;Y)\\
                		&\overset{H_3}{=} I(Y,G_N;G;Z) - I(Z;G;Y)\\
                		&\overset{P_2}{=} I(G_N;G;Z|Y) \\
                		&\overset{P_3}{=} I(G_N;G;Z) - I(G_N;G;Z;Y) \\
                		&\overset{H_2}{=} I(G_N;G;Z) \ \ \ (\text{Note that}\  I(G_N;G;Z;Y) \leq I(G_N;Y))\\
                		&\overset{P_3}{=} I(G_N;Z) - I(G_N;Z|G) \\
                		&\overset{H_1}{=} I(Z;G_N)
            \end{aligned}
            \end{equation}
        Thus we can have $I(Z;G|Y)=0 \iff I(Z;G_N)=0$.
        \end{proof}
    \end{prop}
	
	\begin{lemma}\label{lemma:1}
        $(L_1)$ Let $G$ and $Y$ be random variables with joint distribution $p(G,Y)$, $Z$ is a representation of graph $G$, $S$ denotes the explanatory subgraph, if $Z$ is sufficient for $Y$, then 
        \begin{equation}
            I(S;Y|Z) = 0
        \end{equation}
        
        \noindent Hypothesis:
        
        $(H_1)$ $S$ is an explanation of $G$: $I(S;Y|Z,G)=0$
        
        \noindent Thesis:
        
        $(T_1)$ $I(G;Y|Z)=0 \Rightarrow I(S;Y|Z) = 0$ 
        \begin{proof}
            \begin{equation}
            \begin{aligned}
                I(S;Y|Z) &\overset{P_3}{=} I(S;Y|Z,G) + I(S;Y;G|Z)\\
                           &\overset{H_1}{=} I(S;Y;G|Z) \\
                           &\overset{P_3}{=} I(G;Y|Z) - I(G;Y|Z;S)\\
                           &\leq I(G;Y|Z)
            \end{aligned}
            \end{equation}
        Since $I(S;Y|Z)$ is non-negative,  $I(G;Y|Z)=0 \Rightarrow I(S;Y|Z) = 0$.
        \end{proof}
    \end{lemma}
	
	\begin{lemma}\label{lemma:2}
        $(L_2)$  Let $G$ and $Y$ be random variables with joint distribution $p(G,Y)$, $Z$ is a representation of graph $G$, $S$ denotes the explanatory subgraph, if $Z$ is necessary  for $Y$, then 
        \begin{equation}
            I(S;Z|Y) = 0
        \end{equation}
        
        \noindent Hypothesis:
        
        $(H_1)$ $S$ is an explanation of $G$: $I(S;Z|G,Y)=0$
        
        \noindent Thesis:
        
        $(T_1)$ $I(Z;G|Y)=0 \Rightarrow I(S;Z|Y) = 0$ 
        \begin{proof}
            \begin{equation}
            \begin{aligned}
                I(S;Z|Y) &\overset{P_3}{=} I(S;Z;G|Y) + I(S;Z|G,Y)\\
                           &\overset{H_1}{=} I(S;Z;G|Y) \\
                           &\overset{P_3}{=} I(Z;G|Y) - I(Z;G|S,Y)\\
                           &\leq I(Z;G|Y)
            \end{aligned}
            \end{equation}
        Since $I(S;Z|Y)$ is non-negative,  $I(Z;G|Y)=0 \Rightarrow I(S;Z|Y) = 0$.
        \end{proof}
    \end{lemma}

	\begin{theorem} \label{th:connection}
	Let $G$ and $Y$ be random variables with joint distribution $p(G,Y)$. Given unsupervised graph representation $Z$ and its explanations $S$ in graph $G$. Assume that $S$ is conditionally independent from $Y$ and $Z$ when $G$ is observed, then
	\begin{enumerate}
	\item if $Z$ is sufficient for $Y$, then $I(S;Z) \geq I(S;Y)$;
	\item if $Z$ is necessary for $Y$, then $I(S;Z) \leq I(S;Y)$;
	\item if $Z$ is sufficient and necessary for $Y$, then $I(S;Z) = I(S;Y)$;
	\end{enumerate}
		\noindent Thesis:
		
		$(T_1)$ $I(G;Y|Z)=0 \Rightarrow I(S;Z) \geq I(S;Y)$ 
		
		$(T_2)$ $I(Z;G|Y)=0 \Rightarrow I(S;Z) \leq I(S;Y)$ 
		
		$(T_3)$ $I(G;Y|Z)=0\ and\  I(Z;G|Y)=0 \Rightarrow I(S;Z) = I(S;Y)$ 
		
	\begin{proof}
		We prove the thesis $T_1$ first:
		\begin{equation}
		\begin{aligned}
		I(S;Y) &\overset{P_3}{=} I(S;Z;Y) + I(S;Y|Z) \\
		&\overset{L_1}{=} I(S;Z;Y)  \ \ (\text{see proof in \cref{lemma:1}}) \\
		&\overset{P_3}{=} I(S;Z) - I(S;Z|Y) \\
		&\leq I(S;Z)
		\end{aligned}
		\end{equation}
		
		Then, for thesis $T_2$, we have: 
		\begin{equation}
		\begin{aligned}
			I(S;Y) &\overset{P_3}{=} I(S;Z;Y) + I(S;Y|Z) \\
			&\geq I(S;Z;Y)\\
			&\overset{P_3}{=} I(S;Z) - I(S;Z|Y)\\
			&\overset{L_2}{=} I(S;Z)   \ \ (\text{see proof in \cref{lemma:2}})
		\end{aligned}
	\end{equation}
	
	Finally, having thesis $T_1$ and $T_2$, thesis $T_3$ can be proved simply.

	\end{proof}
	\end{theorem}
    
    \section{Explanation algorithms}
    The algorithm of USIB is shown in \cref{alg:usib}. We first get the unsupervised graph representations by feeding the graph instances into the target model (line 1-2). Then relaxed explanatory subgraphs $\hat{S}$ is generated by the subgraph generator for each batch (line 6-7). After that, we estimate the mutual information and optimize the subgraph generator at the same time (line 8-9). Finally, we get the explanatory subgraph $S$ by selecting edges with top-n edge weights on $\hat{S}$.
    \begin{algorithm}[h]
        \caption{Pseudo-code for USIB} \label{alg:usib}
        {\bf Input:} A set of graphs where $k$-th graph is denoted by $G^{(k)}$, the trade-off parameter $\beta$, the temperature parameter $\tau$, batch size $K$, selection ratio $r$, the target model $f_t$;\\
        {\bf Output:} 
        The explanatory subgraphs $S^{(k)}$;
        \begin{algorithmic}[1]
        \For {each graph $G^{(k)}$}  
            \State $Z^{(k)} = f_t(G^{(k)})$;
        \EndFor
        \State $\mathcal{B}_1, \mathcal{B}_2, \cdots, \mathcal{B}_b \leftarrow $ Batch sample with batch size $K$;
        \For {each epoch}
            \For {each batch $\mathcal{B}_i$}
                \For {$k \in \mathcal{B}_i$}
                    \State $\hat{S}^{(k)} = g_{\theta}(G^{(k)})$;
                \EndFor
                \State Calculate loss with Eq.(10);
                \State Update parameters with Adam optimizer;
            \EndFor
        \EndFor
        \For {each graph $G^{(k)}$} 
            \State  $\hat{S}^{(k)} = g_{\theta}(G^{(k)})$;
            \State $S^{(k)} \leftarrow$ Select top-n edges with selection ratio $r$ on $\hat{S}^{(k)}$;
        \EndFor
        \end{algorithmic}
    \Return The explanatory subgraphs $S^{(k)}$;
    \end{algorithm}
    
	\section{Experiments}
	\subsection{Datasets}
	We summarize the statistics of the datasets in \cref{tab:statistics}. The first three datasets are obtained from Pytorch Geometric Library\footnote{https://pytorch-geometric.readthedocs.io/en/latest/modules/datasets.html} while the last one can be found in work~\cite{wx2021refine}.
	\begin{table*}[ht]
    \centering
	\normalsize
	\setlength{\tabcolsep}{5pt}
	\caption{Statistics of dataset.}\label{tab:statistics}
    \resizebox{0.6\textwidth}{!}{
    \begin{tabular}{lcccccccc} 
    \toprule
    Dataset&Graphs\#&Classes\#&Avg.Nodes\#&Avg.Edges\#\cr
    \midrule
    Mutagenicity&4337&2&30.32&61.54\cr
    NCI1&4110&2&29.87&64.6\cr
    PROTEINS&1113&2&39.06&145.63\cr
    BA3&3000&3&21.92&29.51\cr
    \bottomrule
    \end{tabular}}
\end{table*}

\subsection{Implementation details.}

\paragraph{Target models.} We follow the official implementations of target models \footnote{https://github.com/fanyun-sun/InfoGraph}\footnote{https://github.com/Shen-Lab/GraphCL}\footnote{https://github.com/susheels/adgcl}.  For comparison fairness, we adopt the same encoder architecture for three target models. Specifically, a 3-layer GIN with hidden dimension as 64 is adopted. Output from every GIN layer is concatenated as node representations and the graph-level representations is extracted by conducting add pooling on node representations. Following each linear layer is a ReLU activation function. 

\paragraph{Baselines.}
Here we introduce more details of baseline methods:
\begin{itemize}
    \item \textbf{GNNExplainer}~\cite{ying2019gnnexplainer} optimizes soft masks for edges in the graph, where each mask indicates an edge weight. The soft masks are trained individually for each graph instance.
    \item \textbf{PG-Explainer}~\cite{luo2020parameterized} adopts a parameterized neural network to generate explanations for GNNs. The neural network is trained on multiple instances and thus provides a global view for explanations.
    \item \textbf{PGM-Explainer}~\cite{vu2020pgm} observes the prediction change when the nodes are randomly perturbed, and then a Bayesian network is trained on these observations to capture the dependency among the nodes and the prediction.
    \item \textbf{Refine}~\cite{wx2021refine} follows a pre-training and fine-tuning strategy to generate explanations incorporating multi-granularity information.
    \item \textbf{IG}~\cite{sundararajan2017axiomatic} adopts the integrated gradients as heuristic scores to evaluate the importance of graph edges. Specifically, the integrated gradients are calculated by accumulating gradients with respect to edge weights while interpolating between a baseline and the current instance.
    \item \textbf{SA}~\cite{baldassarre2019explainability} calculates the gradients of the output with respect to edge weights and use the gradients as heuristic scores.
    \item \textbf{GradCam }~\cite{pope2019explainability} calculates the gradients of the output with respect to edge weights first and then multiply the gradients and the edge weights as heuristic scores.
\end{itemize}
To adapt the gradient-based explanation methods, \ie, IG, SA and GradCam to explain unsupervised graph representations, we redesign the heuristic scores to be gradients of $l_2$ norm of graph representations instead of classification probabilities.

\paragraph{USIB.} There are mainly two components in USIB: subgraph generator and mutual information estimator. Specifically,
\begin{itemize}
    \item Subgraph generator: $\text{GNN Type} = \text{ARMAConv}$, $hidden\ dim=32$, $num\ layers=2$, $batch\ normalization = True$, $pooling = add$, $activation = Relu()$,\ $num\ layers\ of\ mlp=2$, $activation\ of\ mlp = Tanh()$, $hidden\ dim\ of\ mlp = 32$, $\tau = 0.1$
    \item Mutual information estimator: $\text{GNN Type} = \text{GCN}$, $hidden\ dim=32$, $num\ layers=3$, $batch\ normalization = True$, $pooling = add$, $activation = Relu()$,\ $num\ layers\ of\ mlp=2$, $activation\ of\ mlp = Relu()$, $hidden\ dim\ of\ mlp = 32$
\end{itemize}
Moreover, the optimization of USIB is performed using Adam and the learning rates is fixed as $0.001$. Batch size of all datasets is set as 64, and number of epochs is 10. The implementation of USIB is based on the Pytorch Geometric library. 
\paragraph{Hardware settings.} All experiments are conducted on a Linux machine with an Nvidia GeForce RTX 3080 GPU with 24GB memory. CUDA version is 11.1 and Driver Version is 460.67.

\begin{figure}[t] 
\centering 
\includegraphics[width=0.99\textwidth]{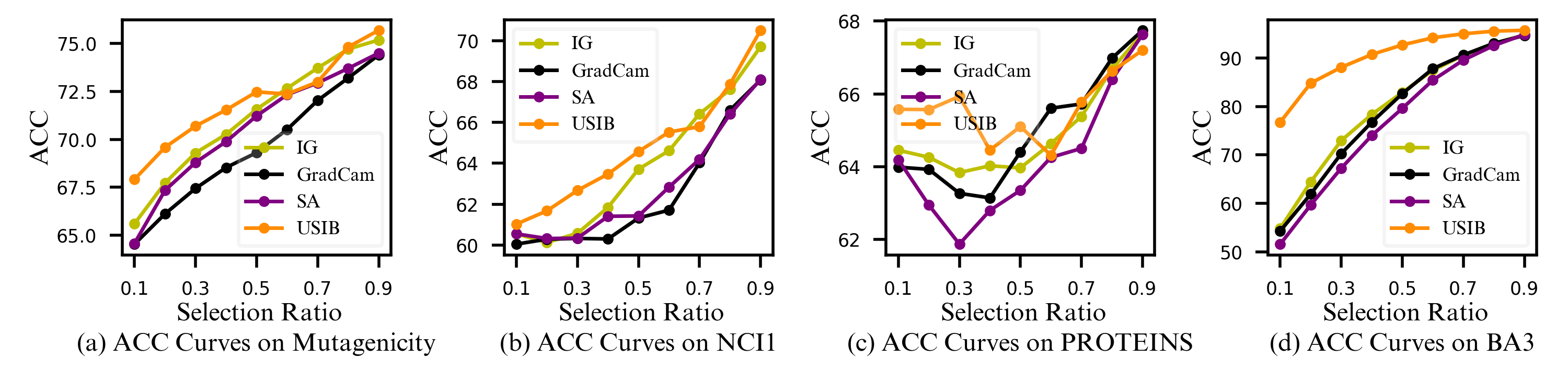} 
\caption{ACC curve for of different explainers on GCL.} 
\label{fig:gcl_acc_curve} 
\end{figure}

\begin{figure}[t] 
\centering 
\includegraphics[width=0.99\textwidth]{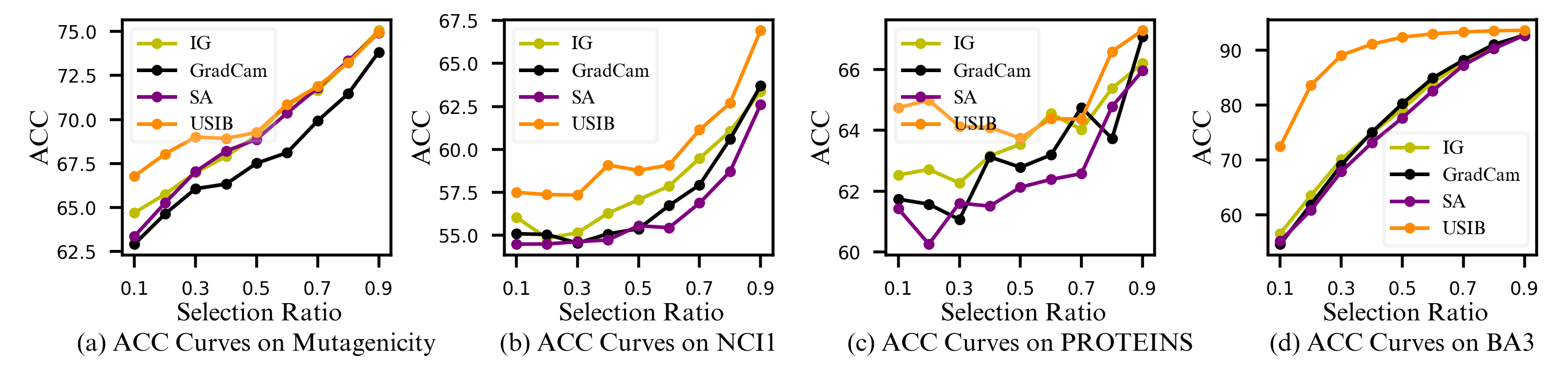} 
\caption{ACC curve for of different explainers on ADGCL.} 
\label{fig:adgcl_acc_curve} 
\end{figure}
\subsection{Experimental results on other target models}
To show the explanation results on GCL and ADGCL, we summarize the ACC-AUC score in \cref{tab:acc_auc_other} and ACC curves in \cref{fig:gcl_acc_curve} and \cref{fig:adgcl_acc_curve}. We can observe similar conclusions from results on GCL and ADGCL: USIB outperforms other explanation methods in unsupervised settings and can be comparable to baseline explanation methods in supervised settings, which is consistent with our analysis on Infograph.

	\begin{table*}[ht]
	\centering
	\normalsize
	\setlength{\tabcolsep}{5pt}
	\caption{Comparison of our USIB and other baseline explainers. The relative improvement is calculated on unsupervised setting methods. \textbf{Bold} indicates the best results.}\label{tab:acc_auc_other}
	\resizebox{0.99\textwidth}{!}{
		\begin{tabular}{lllccccc}
			\toprule
			\multirow{2}{*}{Target Model}&\multirow{2}{*}{Setting}&\multirow{2}{*}{Method}&
			Mutagenicity&NCI1&PROTEINS&\multicolumn{2}{c}{BA3}\cr
			\cmidrule(lr){7-8}
			&&&ACC-AUC&ACC-AUC&ACC-AUC&ACC-AUC&Recall@5\cr
			\midrule
            \multirow{9}{*}{GCL}&\multirow{4}{*}{Supervised}
            &PG-Eplainer&$69.43\pm 2.12$&$63.39\pm 2.00$&$67.08\pm 1.00$&$88.53\pm 2.01$&$27.08\pm 1.98$\cr
            &&GNNExplainer&$71.94\pm 2.20$&$63.87\pm 2.02$&$63.98\pm 1.07$&$77.28\pm 1.80$&$18.63\pm 0.30$\cr
            &&PGM-Explainer&$68.77\pm 0.29$&$62.07\pm 0.52$&$66.07\pm 1.24$&$84.53\pm 1.15$&$24.73\pm 0.07$\cr
            &&ReFine&$70.19\pm 0.76$&$64.25\pm 0.69$&$\bf{67.25\pm 1.46}$&$83.37\pm 2.75$&$20.37\pm 6.10$\cr
            \cmidrule(lr){2-8}&\multirow{5}{*}{Unsupervised}
            &IG&$71.20\pm 0.73$&$63.98\pm 1.14$&$64.99\pm 1.25$&$79.91\pm 2.41$&$14.75\pm 2.94$\cr
            &&GradCam&$69.57\pm 0.98$&$62.52\pm 0.72$&$64.99\pm 0.78$&$79.10\pm 1.21$&$19.49\pm 1.39$\cr
            &&SA&$70.59\pm 0.66$&$62.84\pm 0.78$&$64.27\pm 1.44$&$77.17\pm 1.06$&$16.95\pm 1.92$\cr
            &&USIB&$\bf{72.01\pm 0.89}$&$\bf{64.79\pm 1.50}$&$65.62\pm 2.09$&$\bf{90.37\pm 3.95}$&$\bf{29.50\pm 1.13}$\cr
            \cmidrule(lr){3-8}& &Relative Improvement
            & 1.14\% & 1.27\% & 0.97\%& 13.09\%& 51.36\%\cr
            \midrule
            \multirow{9}{*}{ADGCL}&\multirow{4}{*}{Supervised}
            &PG-Eplainer&$69.85\pm 0.41$&$59.10\pm 0.70$&$66.60\pm 1.29$&$85.53\pm 2.80$&$26.63\pm 1.36$\cr
            &&GNNExplainer&$66.06\pm 0.31$&$54.80\pm 0.57$&$62.93\pm 1.16$&$75.73\pm 2.22$&$18.46\pm 0.09$\cr
            &&PGM-Explainer&$67.26\pm 0.40$&$57.73\pm 0.69$&$64.51\pm 0.89$&$81.97\pm 2.37$&$24.76\pm 0.07$\cr
            &&ReFine&$69.19\pm 0.68$&$\bf{60.58\pm 1.59}$&$\bf{66.84\pm 1.12}$&$84.54\pm 5.77$&$17.30\pm 7.02$\cr
            \cmidrule(lr){2-8}&\multirow{5}{*}{Unsupervised}
            &IG&$69.48\pm 0.96$&$57.91\pm 0.93$&$63.82\pm 0.70$&$77.72\pm 3.02$&$9.76\pm 3.65$\cr
            &&GradCam&$67.87\pm 1.10$&$57.13\pm 1.05$&$63.03\pm 1.05$&$77.51\pm 3.10$&$15.67\pm 1.99$\cr
            &&SA&$69.24\pm 0.93$&$56.40\pm 1.23$&$62.65\pm 1.40$&$76.38\pm 2.85$&$10.68\pm 1.99$\cr
            &&USIB&$\bf{70.33\pm 1.31}$&$59.99\pm 1.65$&$64.93\pm 1.95$&$\bf{89.11\pm 3.71}$&$\bf{29.55\pm 0.88}$\cr
            \cmidrule(lr){3-8}& &Relative Improvement
            & 1.22\% & 3.59\% & 1.92\%& 14\%& 88.58\%\cr
			\bottomrule
		\end{tabular}
	}
\end{table*}
\begin{figure}[ht]
    \centering
    \includegraphics[width=0.99\textwidth]{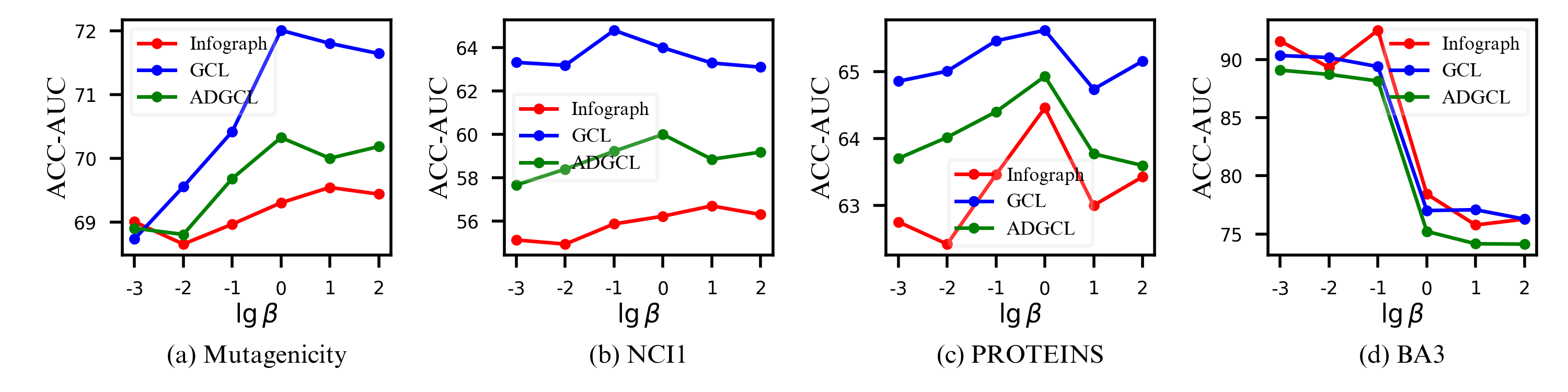}
    \caption{Influence of hyper-parameter $\beta$.}\label{fig:beta_study}
\end{figure}

\begin{table*}[ht]
    \centering
	\normalsize
	\setlength{\tabcolsep}{5pt}
	\caption{Time costs (in second).}\label{tab:time}
    \resizebox{0.5\textwidth}{!}{
    \begin{tabular}{lcccccccc} 
    \toprule
     &Mutagenicity&NCI1&PROTEINS&BA3\cr
    \midrule
    PG-Explainer&8.06&6.38&1.74&4.74\cr
    GNNExplainer&41.97&43.38&11.35&26.87\cr
    PGM-Explainer&570.16&605.76&151.77&287.75\cr
    Refine&7.28&6.81&1.79&4.87\cr
    IG&7.36&6.48&2.23&5.50\cr
    GradCam&0.55&0.49&0.13&0.54\cr
    SA&0.63&0.47&0.15&0.40\cr
    USIB&7.27&6.37&1.82&4.88\cr
    \bottomrule
    \end{tabular}}
\end{table*}

\subsection{Parameter Study}
To investigate the influence of hyper-parameter $\beta$ in our USIB objective, we show the ACC-AUC score with different value of $\beta$ in \cref{fig:beta_study}. We can observe a similar tendency for different target models on various datasets, \ie, either a too low or too high value of $\beta$ degrades the performance of USIB. 
The observation indicates a trade-off between informativeness and compression of explanatory subgraphs. 
Specifically, a higher value of $\beta$ leads to higher compressibility, while the informativeness of explanatory subgraphs is not guaranteed. 
Additionally, the USIB will lead to a trivial solution when the value of $\beta$ is too small, while the edge weights learned for the graph instances will be less informative.

\subsection{Time complexity}
The main time cost during the inference procedure is the calculation of the relaxed edge weights by the neural network. We summarize the time costs in \cref{tab:time}. The time cost of our methods is similar to PG-Explainer as we adopt the same architecture to generate the edge weights. GradCam and SA are the most efficient methods since they only conduct backpropagation for one time. However, the effectiveness of GradCam and SA is not guaranteed according to our above experimental results.

\end{document}